\documentclass[letterpaper, 10pt, conference]{ieeeconf}  

\IEEEoverridecommandlockouts                              

\usepackage{amssymb}  
\usepackage{amsmath}
\usepackage{algorithm}
\usepackage{algpseudocode}
\usepackage[pdftex]{graphicx}
\usepackage{tikz}
\usepackage{tabu}
\usepackage[caption=false]{subfig}

\newtheorem{proposition}{Proposition}

\newtheorem{definition}{Definition}

\usetikzlibrary{shapes,arrows,automata}
\usetikzlibrary{shadows}
\usetikzlibrary{positioning}

\tikzstyle{plain} = [draw=none,fill=none]
\tikzstyle{decision} = [diamond, draw, top color=white, bottom color=blue!30, 
                            draw=blue!50!black!100, drop shadow,
    text width=1.5cm, text badly centered, inner sep=0pt]
\tikzstyle{block} = [rectangle, draw, top color=white, bottom color=blue!30, 
                            draw=blue!50!black!100, drop shadow,
    text width=2.5cm, text centered, rounded corners]
\tikzstyle{nodepolicy} = [circle, draw, top color=white, bottom color=blue!30, 
                            draw=blue!50!black!100, drop shadow,
    text width=.8cm, text centered]
\tikzstyle{inputpolicy} = [circle, draw, top color=white, bottom color=green!30, 
                            draw=green!50!black!100, drop shadow,
    text width=.8cm, text centered]
\tikzstyle{outputpolicy} = [circle,trapezium left angle=70,trapezium right angle=-70, draw, top color=white, bottom color=red!30, 
                            draw=red!50!black!100, drop shadow,
    text width=.8cm, text centered]
\tikzstyle{line} = [draw, -latex, ultra thick]
\tikzstyle{cloud} = [draw, ellipse,fill=red!20, node distance=3cm,
    minimum height=2em]

\newcommand{\mbs}[1]{\ensuremath{\boldsymbol{#1}}}

\newcommand{\thetav}{\mbs{\theta}}

\newcommand{\kv}{\mathbf{k}}

\newcommand{\x}{\mathbf{x}}
\newcommand{\y}{\mathbf{y}}

\newcommand{\I}{\mathbf{I}}
\newcommand{\K}{\mathbf{K}}

\newcommand{\X}{\mathbf{X}}

\newcommand{\w}{\mathbf{w}}

\newcommand{\N}{\mathcal{N}}

\renewcommand{\baselinestretch}{0.98}

\title{\LARGE \bf
Bayesian Optimization with Adaptive Kernels for Robot Control
}

\author{Ruben Martinez-Cantin
\thanks{*This work was supported in part by projects DPI2015-65962-R (MINECO/FEDER, UE), CUD2013-05 and CUD2016-17.}
\thanks{R. Martinez-Cantin is at Centro Universitario de la Defensa, Zaragoza, Spain. 
and SigOpt, Inc. {\tt rmcantin@unizar.es} }
}

\begin{document}

\bstctlcite{IEEEexample:BSTcontrol}

\maketitle
\thispagestyle{empty}
\pagestyle{empty}

\begin{abstract}
Active policy search combines the trial-and-error methodology from policy search with Bayesian optimization to actively find the optimal policy. First, policy search is a type of reinforcement learning which has become very popular for robot control, for its ability to deal with complex continuous state and action spaces. Second, Bayesian optimization is a sample efficient global optimization method that uses a surrogate model, like a Gaussian process, and optimal decision making to carefully select each sample during the optimization process. Sample efficiency is of paramount importance when each trial involves the real robot, expensive Monte Carlo runs, or a complex simulator. Black-box Bayesian optimization generally assumes a cost function from a stationary process, because nonstationary modeling is usually based on prior knowledge. However, many control problems are inherently nonstationary due to their failure conditions, terminal states and other abrupt effects. In this paper, we present a kernel function specially designed for Bayesian optimization, that allows nonstationary modeling without prior knowledge, using an adaptive local region. The new kernel results in an improved local search (exploitation), without penalizing the global search (exploration), as shown experimentally in well-known optimization benchmarks and robot control scenarios. We finally show its potential for the design of the wing shape of a UAV.
\end{abstract}

\section{INTRODUCTION}

When a baby is learning a new behavior, like walking or grasping, she performs it several times, trying to improve the outcome of the behavior in each trial or exploring new strategies. This trial and error methodology share many points with the model-free direct policy search methodology that has been quite successful in robotics \cite{deisenroth2013survey}. These algorithms are designed to maximize or minimize a respective reward or cost function as a function of a parametrization of the policy or behavior that the agent is following. Traditionally, the most popular methods were variations of policy gradient methods \cite{Peters_PIICIRS_2006}. Recently, there has been an increasing number of methods that have been applying Bayesian optimization in the context of policy search for robotics. The advantages of Bayesian optimization for policy search are: a) ability to find the global optimum, b) after testing few policies, c) without gradient information and d) without model or instant reward. Recent studies have found connections between Bayesian optimization and the way biological systems adapt and search in nature \cite{Cully2015}.

Bayesian optimization is a classic global optimization method \cite{Mockus78} which has become quite popular recently for being sample efficient \cite{Jones:1998} and applied with great success for machine learning applications \cite{Snoek2012}, 
etc. In the context of robotics it has been applied for robot planning \cite{MartinezCantin09AR,marchant2014bayesian}, control \cite{Calandra2015a,Tesch_2011_7370}, task optimization \cite{KroemerJRAS_66360}, grasping \cite{ubonogueira}, model-free reinforcement learning \cite{MartinezCantin07RSS,Kuindersma01062013}, model-based reinforcement learning \cite{JMLR:v15:wilson14a}, sensor networks \cite{Srinivas10}, etc.

Bayesian optimization is the combination of two main components: a surrogate model which captures all prior and observed information and a decision process which performs the optimal action, i.e.: where to sample next, based on the previous model. Thus, the quality of the surrogate model is of paramount importance as it also affects the optimality of the decision process. Earliest versions of Bayesian optimization used Wiener processes \cite{Mockus78} as surrogate models. It was the seminal paper of Jones et al. \cite{Jones:1998} that introduced the use of Gaussian processes (GP). The GP model is the most popular due to its accuracy, robustness and flexibility, because Bayesian optimization is mainly used in black or grey-box scenarios. The range of applicability of a GP is defined by its kernel function, which sets the family of functions that is able to represent through the reproducing kernel Hilbert space (RKHS) \cite{Rasmussen:2006}. 
From a practical point of view, the standard procedure is to select a generic kernel function, such as the Gaussian (square exponential) or Mat{\'e}rn kernels, and estimate the kernel hyperparameters from data. One property of these kernels is that they are stationary. Although it might be a reasonable assumption in a black box setup, we show in Section \ref{sec:nst} that this reduces the efficiency of Bayesian optimization in most situations. It also limits the potential range of applications. Moreover, nonstationay methods usually require extra knowledge of the function (e.g.: the global trend or the space partition). Being global properties, gathering this knowledge from data requires global sampling, which is contrary to the Bayesian optimization methodology.

The main contribution of the paper is a new set of adaptive kernels for Gaussian processes that are specifically designed to model functions from nonstationary processes but focused on the local region near the optimum. Thus, the new model maintains the philosophy of global exploration/local exploitation. This idea results in an improved sample efficiency of any Bayesian optimization based on Gaussian processes. We call this new method \emph{Spartan Bayesian Optimization} (SBO). The algorithm has been extensively evaluated in many scenarios and applications. Besides some standard optimization benchmarks, we show the applicability to optimal policy learning in reinforcement learning scenarios. Furthermore, to highlight the general purpose of the method, we also show an application of autonomous wing design of a UAV. 

\section{Bayesian optimization with Gaussian processes}
\label{sec:bo}

Consider the problem of finding the minimum of an unknown real valued function $f:\mathbb{X} \rightarrow \mathbb{R}$, where $\mathbb{X}$ is a compact space, $\mathbb{X} \subset \mathbb{R}^d, d \geq 1$. In order to find the minimum, the algorithm has a maximum budget of $N$ evaluations of the target function $f$. The purpose of the algorithm is to select the best query points at each iteration such that the \emph{optimization gap} or \emph{regret} is minimum for the available budget. 

The surrogate model is a Gaussian process $\mathcal{GP}(\x|\mu,\sigma^2,\thetav)$ with inputs $\x \in \mathbb{X}$, scalar outputs $y \in \mathbb{R}$ and an associate kernel or covariance function $k(\cdot,\cdot)$ with hyperparameters $\thetav$. The hyperparameters are estimated from data using Markov Chain Monte Carlo (MCMC) resulting in $m$ samples $\mathbf{\Theta} = \{\thetav_i\}_{i=1}^m$.  
Given a dataset at step $n$ of query points $\X = \{\x_{1:n}\}$ and its respective outcomes $\y = \{y_{1:n}\}$, then the prediction of the Gaussian process at a new query point $\x_q$, with kernel $k_i$ conditioned on the $i$-th hyperparameter sample $k_i = k(\cdot,\cdot|\thetav_i)$ is a normal distribution such as $y_q \sim 1/m \; \sum_{i=1}^m \N(\mu_i,\sigma^2_i|\x_q)$ where:
\begin{equation}
 \begin{split}
\label{eq:predgp}
\mu_i(\x_q) &= \kv_i(\x_q,\X) \K_i(\X,\X )^{-1} \y  \\
\sigma^2_i(\x_q) &= k_i(\x_q,\x_q)\\
&\phantom{=} - \kv_i(\x_q,\X) \K_i(\X,\X )^{-1} \kv_i(\X,\x_q)    
  \end{split}
\end{equation}
being $\kv_i(\x_q,\X)$ the corresponding cross-correlation vector of the query point $\x_q$ with respect to the dataset $\X$
\[
\kv_i(\x_q,\X) = \left[  k_i(\x_q,\x_1), \ldots,  k_i(\x_q,\x_n) \right]^T
\]
and $\K_i(\X,\X) = \K_G + \sigma^2_n\I$ is the Gram matrix $\K_G$ corresponding to kernel $k_i$ for the dataset $\X$, and $\sigma^2_n$ is a noise term to represent stochastic functions or surrogate mismodeling. The prediction at any point $\x$ is a mixture of Gaussians because we use a sampling distribution of $\thetav$.

For the decision process, first we rely on an initial design of $p$ points based on \emph{Latin Hypercube Sampling} (LHS) \cite{Jones:1998}, to avoid initialization bias. Subsequent points are selected using the \emph{expected improvement} criterion \cite{Mockus78} which is defined as the expectation of the improvement function $I(\x) = \max(0,\rho - f(\x))$. This improvement is defined over a incumbent target $\rho$, which in many applications is considered to be the \emph{best outcome} until that iteration $\rho = y_{best}$. Taking the expectation over the mixture of Gaussians of the predictive distribution, we can compute the expected improvement as:
\begin{equation}
  \label{eq:eigen}
  EI(\x) = \sum_{i=1}^m \left[\left(\rho - \mu_i(\x)\right) \Phi(z_i) + \sigma_i(\x) \phi(z_i)\right]
\end{equation}
where $\phi$ and $\Phi$ are the corresponding Gaussian probability density function (PDF) and cumulative density function (CDF), being $z_i = (\rho - \mu_i(\x))/\sigma_i(\x)$. At iteration $n$, we select the next query at the point that maximizes the expected improvement $\x_n = \arg \max_{\x}   EI(\x)$

\subsection{Kernels for Bayesian optimization}
\label{sec:kernels}

Many applications of Gaussian process regression, including Bayesian optimization, are based on the assumption that the process is stationary. This is a reasonable assumption for black-box optimization as it does not assume any extra information on the evolution of the function in the space. For example, the use of the squared exponential (SE) kernel in GPs is quite frequent:
$k_{SE}(\x,\x') = \exp\left(-\frac{1}{2} r^2\right)$ where $r$ is the weighted $L^2$ norm between $\x$ and $\x'$. The hyperparameters of the kernel are the components of the weighting matrix of the norm $\Lambda$. In many applications, the matrix is a simple constant $\Lambda = \theta^{-1} \I$ (isotropic process). If we use a diagonal matrix $\Lambda = diag(\theta_1^{-1},\ldots,\theta_d^{-1})$ (anisotropic process), it is called \emph{automatic relevance determination} (ARD).

Because the SE kernel is infinitely differentiable, it tends to over-smooth functions. For Bayesian optimization, a more suitable kernel is the Mat{\'e}rn kernel family, specifically the the Mat{\'e}rn kernel with $\nu=5/2$, as it provides a good trade off of differentiability/smoothness:
\begin{equation}
  k_{M5}(\x,\x') = \exp\left(-\sqrt{5} r\right) \left(1+\sqrt{5} r + \frac{5}{3} r^2\right)   \label{eq:matern}  
\end{equation}

For these kernels, the hyperparameters $\thetav_l$ represent the length-scales that captures the smoothness or variability of the function in the corresponding dimension \cite{Rasmussen:2006}. Small values of $\thetav_l$ will be more suitable to capture signals with high frequency components; while large values of $\thetav_l$ result in a model for low frequency signals or flat functions. This effect is very important in Bayesian optimization. For the same distance between points, a kernel with smaller length-scale will result in higher predictive variance, therefore the exploration will be more aggressive. This idea was previously explored in Wang et al. \cite{ZiyuWang2016short} by forcing smaller scale parameters to improve the exploration. More formally:
\begin{proposition} \cite{ZiyuWang2016short}
Given two kernels $k_l$ and $k_s$ with large and small length scale hyperparameters respectively, any function $f$ in the RKHS characterized by a kernel $k_l$ is also an element of the RKHS characterized by $k_s$.
\end{proposition}
Thus, using $k_s$ instead of $k_l$ is safer in terms of guaranteeing convergence. However, if the small kernel is used everywhere, it might result in unnecessary sampling of smooth areas.

\subsection{Nonstationary Gaussian processes}
\label{sec:nst}

Consider the problems of Section \ref{sec:walker} where a biped robot (agent) is trying to learn the walking pattern (policy) that maximizes the walking speed (reward). In this setup, there are some policies that reach undesirable states or result in a failure condition, like the robot falling or losing the upright posture. Then, the system returns a null reward or arbitrary penalty. In cases where finding a stable policy is difficult, the reward function may end up being almost flat, except for a small region of successful policies where the reward is actually informative in order to maximize the speed.

Modeling these kind of functions with Gaussian processes require kernels with different length scales for the flat/non-flat regions or specially designed kernels to capture that behavior. Furthermore, Bayesian optimization is inherently a local stationary process depending on the acquisition function. It has a dual behavior of global exploration and local exploitation. Ideally, both samples and uncertainty estimation end up being distributed unevenly, with many samples and small uncertainty near the local optima and sparse samples and large uncertainty everywhere else. 

\begin{definition}
Let $f: \mathbb{R}^d \rightarrow \mathbb{R}$ be a function and $\mathcal{H}_k$ be the reproducing kernel Hilbert space generated by kernel $k(\cdot,\cdot)$.
\begin{itemize}
\item  We say that a function $f(\x)$ is \emph{stationary} if $\exists \; k(\x,\x') = k(\mathbf{\tau})$ where $\mathbf{\tau} = \x-\x'$ and $f\in \mathcal{H}_k$.
\item  In contrast, we say that a function $f(\x)$ is \emph{nonstationary} if $\nexists \; k(\x,\x') = k(\mathbf{\tau})$ where $\mathbf{\tau} = \x-\x'$ and $f \in \mathcal{H}_k$.
\item  Finally, we say that a function $f(\x)$ is \emph{local stationary} if there is a subset $\mathcal{X} \subset \mathbb{R}^d$ so that the function is stationary $\forall \x \in \mathcal{X}$ and nonstationary $\forall \x \in \mathbb{R}^d \setminus \mathcal{X}$.
\end{itemize}
\end{definition}
According to the previous definition, most applications of Bayesian optimization are nonstationary or local stationary. Also, even for stationary problems, we might want different levels of exploration in different regions, which might require using two or more length-scales as seeing in Section \ref{sec:kernels}.

There have been several attempts to model nonstationarity in Bayesian optimization. Bayesian treed GPs were used in Bayesian optimization combined with an auxiliary local optimizer \cite{taddy2009bayesian}. An alternative is to project the input space through a warping function to a stationary latent space \cite{snoek-etal-2014a}. Later, Assael et al. \cite{Assael2014} built treed GPs where the warping model was used in the leaves. These methods were direct application of regression methods, that is, they model the nonstationary property in a global way. However, as pointed out before, sampling in Bayesian optimization is uneven, thus the global model might end up being inaccurate.


\section{Spartan Bayesian Optimization}

Our approach to nonstationarity is based on the model presented in Krause \& Guestrin \cite{krause07nonmyopic} where the input space is partitioned in different regions such that the resulting GP is the linear combination of local GPs: $\xi(\x) = \sum_j \lambda_j (\x) \xi_j(\x)$. Each local GP has its own specific hyperparameters, making the final GP nonstationary even when the local GPs are stationary. In order to achieve smooth interpolation between regions, the authors suggest the use of a weighting function $\omega_j(\x)$ for each region, having the maximum in region $j$ and decreasing its value with distance to region $j$ \cite{krause07nonmyopic}. Then, we can set $\lambda_j(\x) = \sqrt{\omega_j(\x)/\sum_p \omega_p(\x)}$. In practice, the mixed GP can be obtained by a combined kernel function of the form: $k(\x,\x'| \thetav) = \sum_{j} \lambda_j (\x) \lambda_j (\x') k_j(\x,\x'|\thetav)$. A related approach of additive GPs was used by Kandasamy et al. \cite{kandasamy2015high} for Bayesian optimization of high dimensional functions under the assumption that the actual function is a combination of lower dimensional functions.

\begin{figure}
  \centering
  \includegraphics[width=0.6\linewidth]{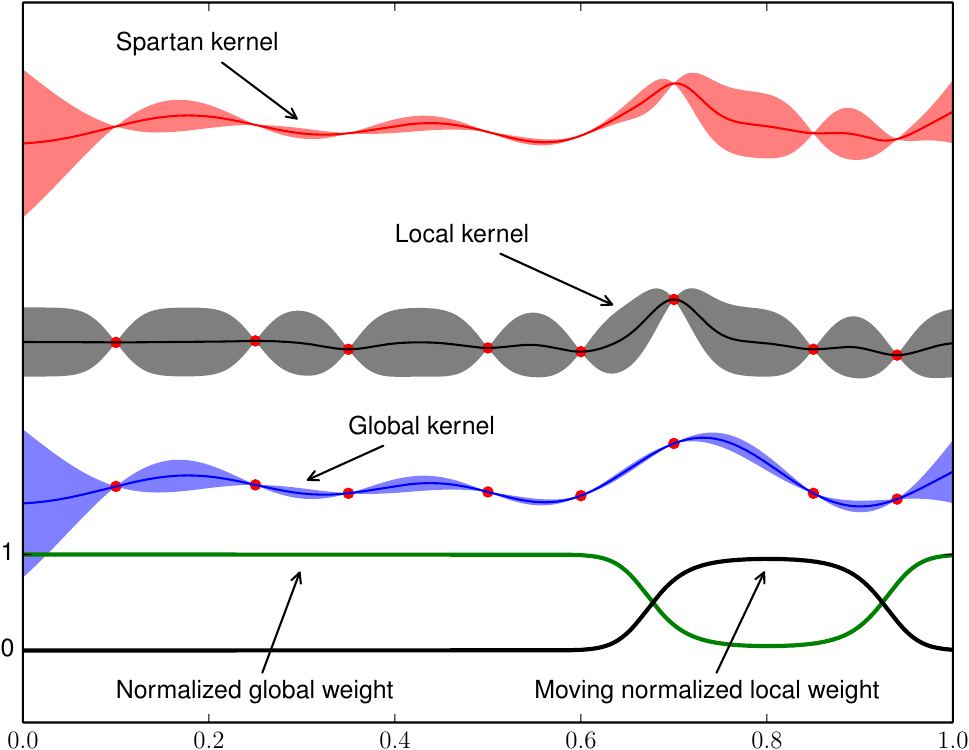}
  \caption{Representation of the Spartan kernel in SBO. Typically, the local and global kernels have a small and large length-scale respectively. The influence of each kernel is represented by the normalized weight at the bottom of the plot. Note how the kernel with small length-scale produces larger uncertainties which is an advantage for fast exploitation, but it can perform poorly for global exploration as it tends to sample equally almost everywhere. On the other hand, the kernel with large length-scale provides a better global estimate, but it can be too constrained locally.}
  \label{fig:domains}
\end{figure}

For Bayesian optimization, we propose the combination of a local and a global kernels and with multivariate normal distributions as weighting functions. We have called this kernel, the Spartan kernel:
\begin{equation}
  \label{eq:spartan}
  \begin{split}
k(\x,\x'| \thetav^{S}) &= \lambda^{(g)}(\x) \lambda^{(g)}(\x') k^{(g)}(\x,\x'| \thetav^g) \\ &+  \lambda^{(l)}(\x| \thetav^{p}) \lambda^{(l)}(\x'| \thetav^{p}) k^{(l)}(\x,\x'| \thetav^l)
  \end{split}
\end{equation}
where the normalized local weight $\lambda^{(l)}(\x| \thetav^{p})$ includes parameters to move the influence region. The unnormalized weights $\omega$ are defined as:
\begin{equation}
  \label{eq:weights}
  \begin{aligned}
    \omega^{(g)} &= \N\left(\psi, \I \sigma^2_{g} \right) \\
    \omega^{(l)} &=  \N\left(\thetav^{p}, \I \sigma^2_{l} \right)
  \end{aligned}
\end{equation}
where $\psi$ and $\thetav^{p}$ can be seen as the centers of the influence region of each kernel while $\sigma^2_{g}$ and $\sigma^2_{l}$ represents the area of influence. The Spartan kernel is shown in Figure \ref{fig:domains}.

\subsubsection{Global weight parameters}
Unless we have prior knowledge of the function, the parameters of the global weight are mostly irrelevant. In most applications, we can use a uniform weight, which can be easily approximated with a large $\sigma^2_{g}$. For example, assuming a normalized input space $\mathcal{X} = [0,1]^d$, we can set $\psi = [0.5]^d$ and $\sigma^2_{g} = 10$.

\subsubsection{Local weight parameters}
For the local weight, we propose to consider the center of the influence area of the local kernel $\thetav^p$ as part of the hyperparameters for the Spartan kernel, that also includes the parameters of the local and global kernels, that is: 
\[
\thetav^{S}=[\thetav^{g}, \thetav^{l}_1,\ldots,\thetav^{l}_M, \thetav^{p}]
\]
Thus, $\thetav^p$ is also estimated from data. The variance of the weight of the local kernel (extension of the influence area) could also be adapted including the terms $\sigma^2$ as part of the kernel hyperparameters $\thetav^{S}$. However, in that case the problem becomes ill-posed, resulting in overfitting. Instead of adding regularization terms, we found simpler to fix the value of $\sigma_l^2$ or fix the number of samples within $2\sigma_l^2$. The second method has the advantage that, while doing exploitation, as the number of local samples increase, the area gets narrower, allowing better local modeling.

\subsubsection{Learning the parameters}
\begin{figure}
  \centering
  \includegraphics[width=0.50\linewidth]{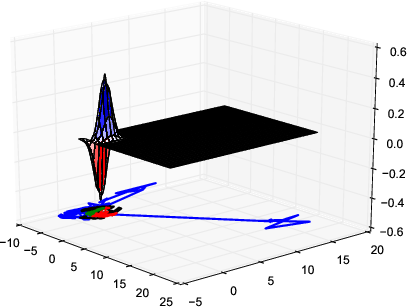}
  \caption{Gramacy function \cite{Assael2014}. The path bellow the surface represents the location of the local kernel as being sampled by MCMC for each BO iteration. Clearly, it moves towards the nonstationary section of the function. For visualization, the path is colored depending on the iteration (start $\rightarrow$ blue $\rightarrow$ black $\rightarrow$ green $\rightarrow$ red $\rightarrow$ end).}
  \label{fig:exp2dres}
\end{figure}

As commented in Section \ref{sec:bo}, when new data is available, all the parameters are updated using MCMC. Therefore, the position of the local kernel $\thetav^p$ is moved each iteration to represent the posterior, as can be seen in Figure \ref{fig:exp2dres}. Due to the sampling behavior in Bayesian optimization, we found that it intrinsically moves more likely towards the more densely sampled areas in many problems, which corresponds to the location of the function minima. Furthermore, as we have $m$ MCMC samples, there are $m$ different positions for the local kernel $\mathbf{\Theta}^p = \{\thetav^p_i\}_{i=1}^m$. 

It is important to note that, although we have described SBO relying on GPs and EI, the Spartan kernel also works with other popular models such as Student-t processes, variational GPs; other criteria such as \emph{upper confidence bound} \cite{Srinivas10}, \emph{relative entropy} \cite{NIPS2014_pesshort}; and specific configuration such as trajectory aware kernels \cite{JMLR:v15:wilson14a,marchant2014bayesian}. 

The intuition behind SBO is the same of the sampling strategies in Bayesian optimization: \emph{the aim of the model is not to approximate the target function precisely in every point, but to provide information about the location of the minimum}. Many optimization problems are difficult due to the fact that the region near the minimum is heteroscedastic, i.e.: it has higher variability than the rest of the space, like the function in Figure \ref{fig:exp2dres}. In this case, SBO greatly improves the performance of the state of the art in Bayesian optimization. 

\section{Active policy search}
Reinforcement learning algorithms usually rely on variants of the Bellman equation to optimize the policy step by step considering each instantaneous reward $r_t$ separately. Some algorithms also rely on partial or total knowledge of the transition model. Other methods tackle the optimization problem directly, considering the problem of finding the optimal policy as a stochastic optimization problem, being called \emph{direct policy search}. In that way, the use of Bayesian optimization for reinforcement learning falls in the family of direct policy search, being called \emph{active policy search} \cite{MartinezCantin07RSS} for its connection with active learning and how samples are carefully selected based on current information.

The main advantage of using Bayesian optimization to compute the optimal policy is that it can be done with very little information. In fact, as soon as we are able to simulate scenarios and return the total reward $\sum_{t=1}^Tr_t$, we do not need to access the dynamics, the instantaneous reward or the current state of the system. Furthermore, there is no need for space or action discretization, building complex features or \emph{tile coding} \cite{sutton1998}. We found that for many control problems, a simple, low dimensional policy is able to achieve state-of-the-art performance if properly optimized. We also solve stochasticity by running each episode several times and returning the average reward, as an approximation of the expected reward.

A frequent issue for applying general purpose optimization algorithms for policy search is that, in many problems, the occurrence of \emph{failure} states or scenarios results in large discontinuities or flat regions due to large penalties for all failing policies. This is opposed to the behavior of the reward near the optimal policy where small variations on a suboptimal policy can considerably change the performance achieved. Therefore, the resulting reward function presents a nonstationary behavior with respect to the policy.

\section{Evaluation and results}
\label{sec:results}

We have selected a variety of benchmarks from the optimization, RL/control and robotics literature. For evaluation purposes and to highlight the robustness of SBO, we took the simpler approach to fix the variance of $\omega_l$. We found that a value of $\sigma^2_{l} = 0.05$ was robust enough in all the experiments once the input space was normalized to the unit hypercube.

\begin{figure*}
  \centering
  \subfloat{\includegraphics[width=0.24\textwidth]{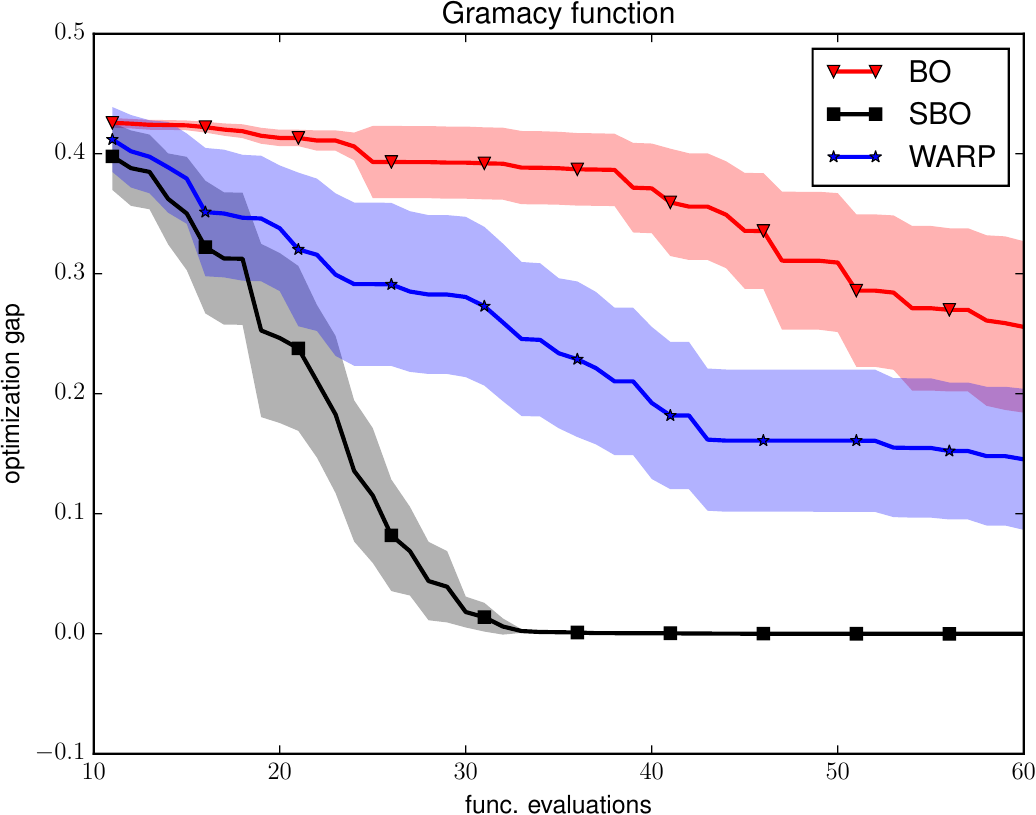} \label{sfig:exp2d}}
  \subfloat{\includegraphics[width=0.24\textwidth]{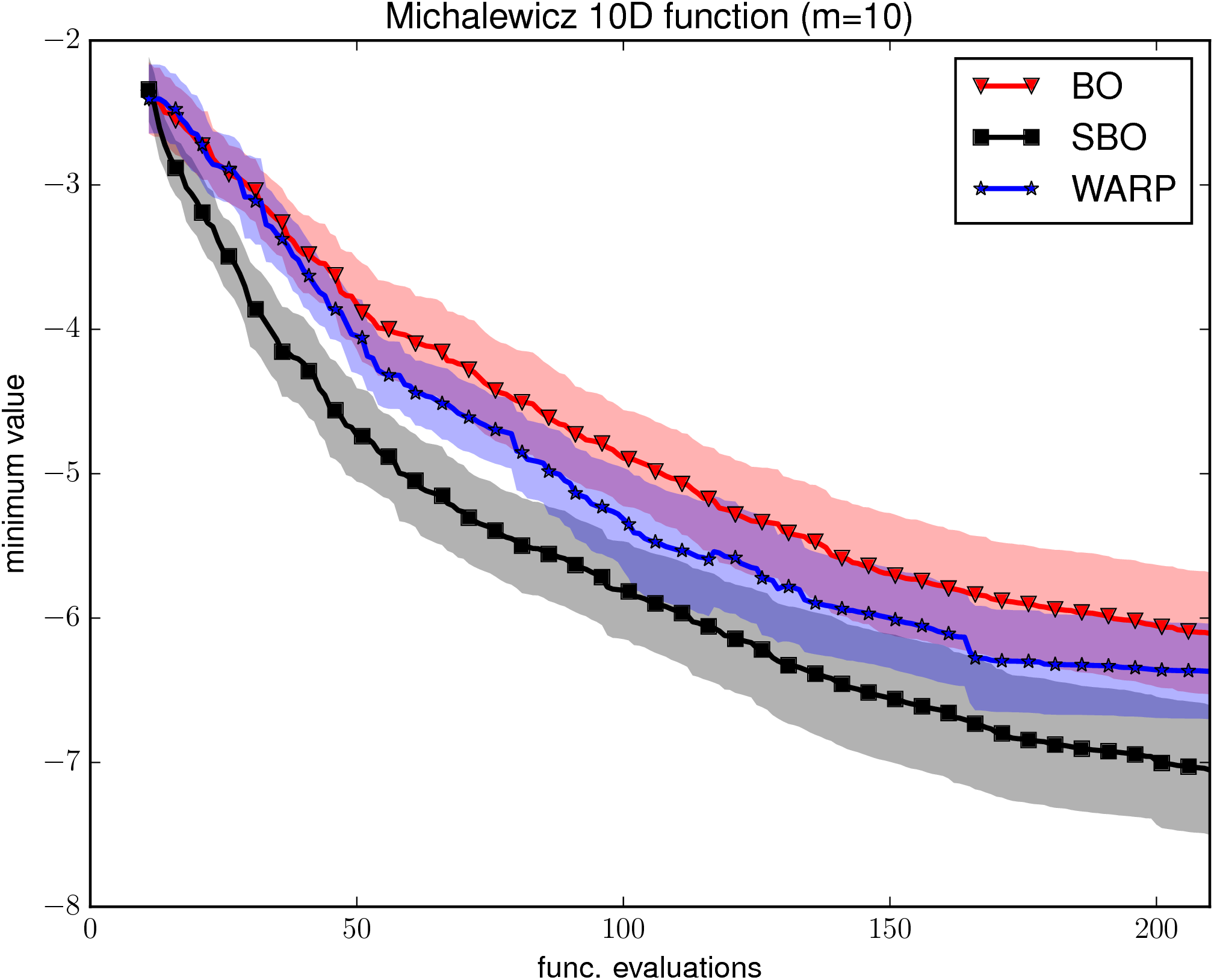} \label{sfig:micha}}
  \subfloat{\includegraphics[width=0.24\textwidth]{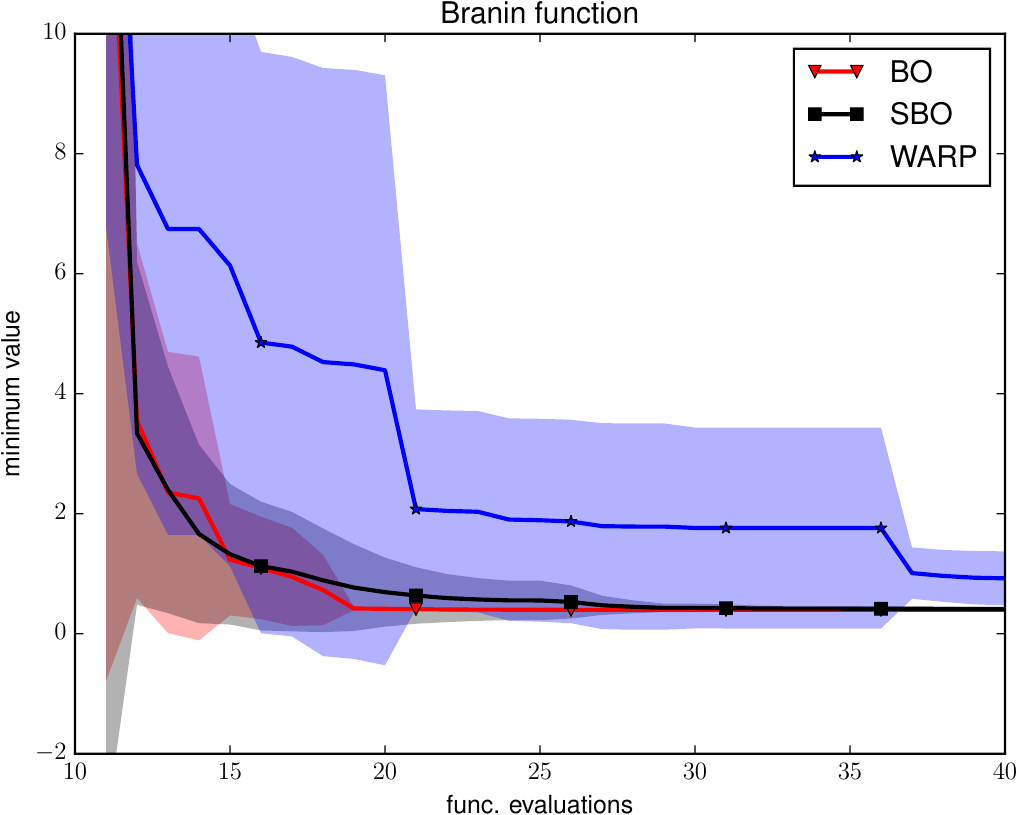} \label{sfig:branin}}
  \subfloat{\includegraphics[width=0.24\textwidth]{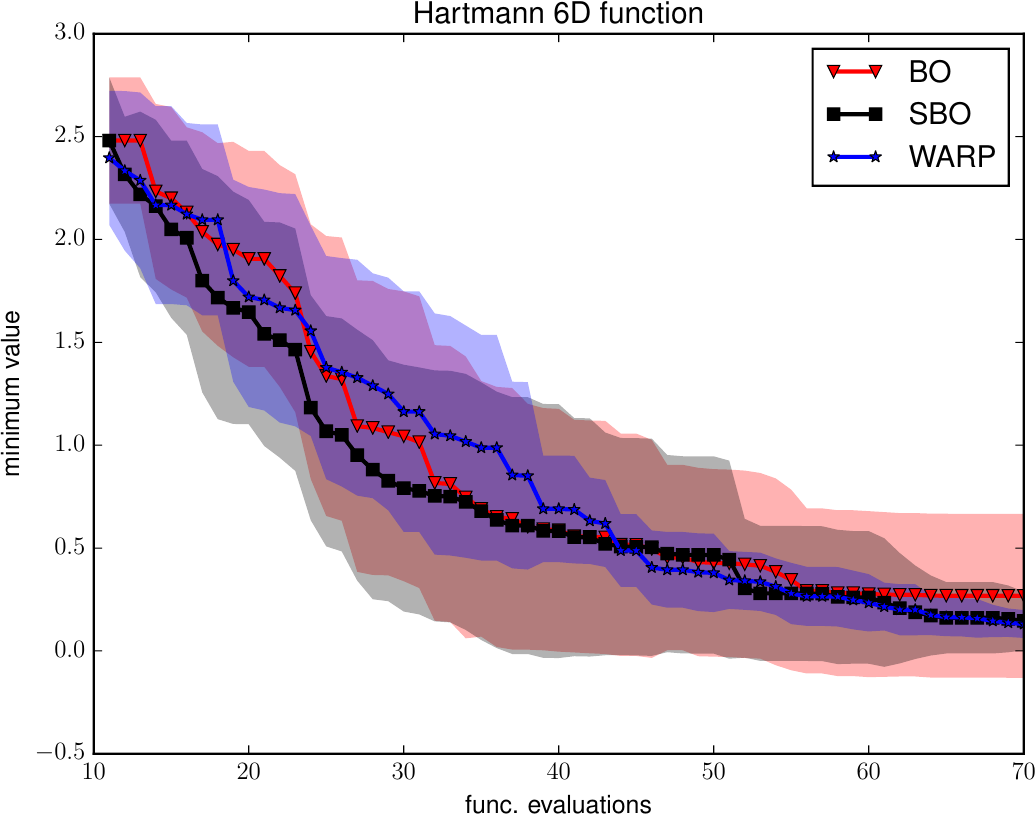} \label{sfig:hart}}
  \caption{a) Gramacy function. b) Michalewicz 10D function with m=10. c) Branin-Hoo function, d) Hartmann 6D function. For the nonstationary functions, a) and b), the proposed SBO method results in an outstanding convergence speed compared to the state of the art. For the Gramacy function, SBO finds the minimum in about 30 function evaluations in all tests. For the stationaty functions, c) and d) BO and SBO are barely identical, with SBO producing more accurate results and with smaller uncertainty. The WARP method sometimes improves over standard BO (a,b and d) or produces worse results (c).}
  \label{fig:optbench}
\end{figure*}

Although this method allows for any combination of local and global kernels, for the purpose of evaluation, we used the Mat{\'e}rn kernel from equation (\ref{eq:matern}) with ARD for both --local and global-- kernels. Furthermore, the length-scales were initialized with the same prior for the both kernels. Therefore, we let the data determine which kernel has smaller length-scale. We found that the typical result is the behavior from Figure \ref{fig:domains}. However, in some problems, the method may learn a model where the local kernel has a larger length-scale (i.e.: smoother and smaller variance) than the global kernel, which may also improve the convergence in plateau-like functions. Besides, if the target function is stationary, the system might end up learning a similar length-scale for both kernels, thus being equivalent to a single kernel. We can say that standard BO is a special case of SBO where the local and global kernels are the same. 

Given that for a single Mat{\'e}rn kernel with ARD, the number of kernel hyperparameters is the dimensionality of the problem, $d$, the number of hyperparameters for the Spartan kernel in this setup is $3d$. As we will see in the experiments, this is the only drawback of SBO compared to standard BO, as it requires running MCMC in a larger dimensional space, which results in higher computational cost. However, because SBO is more efficient, the extra computational cost can be easily compensated by a reduced number of samples.


We implemented Spartan Bayesian Optimization (SBO) using the BayesOpt library \cite{MartinezCantin14jmlr}. This allowed us to evaluate the setup for many surrogates and criteria. For comparison, we also implemented the input warping (WARP) method from Snoek et al. \cite{snoek-etal-2014a}. To our knowledge, this is the only Bayesian optimization algorithm that has deal with nonstationarity using GPs in a fully correlated way. 

For the experiments reported here we used: a Gaussian process with unit mean function like in Jones et al. \cite{Jones:1998}. For BO and WARP we also used a Mat{\'e}rn kernel $\nu=5/2$ with ARD. The kernel hyperparameters, including $\thetav^p$ for SBO and $(\alpha,\beta)$ for the warping functions were estimated using MCMC (i.e.: slice sampling). Due to the computational burden of MCMC, we used a small number of samples (i.e.: 10), while trying to decorrelate every resample with large burn-in periods (i.e.: 100 samples) as in Snoek et al. \cite{Snoek2012}. All experiments were repeated 20 times using common random numbers. The starting function evaluations of the plots represents the initial design using \emph{latin hypercube sampling}. Plots show the average results over all runs with 95\% confidence intervals.

\subsection{Optimization benchmarks}

We evaluated the algorithms on a set of well-known test functions for global optimization both smooth or with sharp drops (see Figure \ref{fig:optbench}). The selected functions have become a standard in the Bayesian optimization literature. 

First, we show the results for the optimization benchmarks for functions with sharp drops where local optimization is fundamental. Figure \ref{sfig:exp2d} shows the results of the Gramacy function found in \cite{Assael2014}. Our method (SBO) provides excellent results, by reaching the optimum in less than 35 samples for all tests. Because the function is nonstationary, the WARP method outperforms standard BO, but its convergence is much slower than SBO. Figure \ref{sfig:micha} shows the results for the Michalewicz function. This function has a parameter to define the dimensionality $d$ and the steepness $m$. This function is known to be a hard benchmarks in global optimization due to the many local minima ($d!$) and steep drops. We used $d=10$ and $m=10$, resulting in $3628800$ minima with very steep edges. For this problem, SBO clearly outperforms the rest of the methods by a large margin.

We have also evaluated stationary and smooth functions with large valleys near the global minimum.  Bayesian optimization is more suitable for these functions and standard kernels perform well in general. Therefore, there is barely room from improvement. However, we show that, even in this situation, SBO is equal or better than standard BO. In terms of accuracy, there is no penalty for the extra complexity of the SBO model, while the WARP method may require more samples due to the extra complexity. For the Branin-Hoo function (see Figure \ref{sfig:branin}), the differences between SBO and BO are insignificant. Meanwhile, the warping function in WARP introduces an exploration bias at early stages, resulting in slower convergence. For the Hartmann 6D function (see Figure \ref{sfig:hart}), the differences are small, which imply that the function is most likely stationary and simple to exploit. However, we can see that during the final iterations, nonstationary methods (SBO and WARP) slightly improve standard BO, both in terms of average result (convergence) and variance (robustness).

\subsection{Reinforcement learning experiments}
\label{sec:reinf-learn}

\begin{figure*}
  \centering
\includegraphics[width=0.25\linewidth]{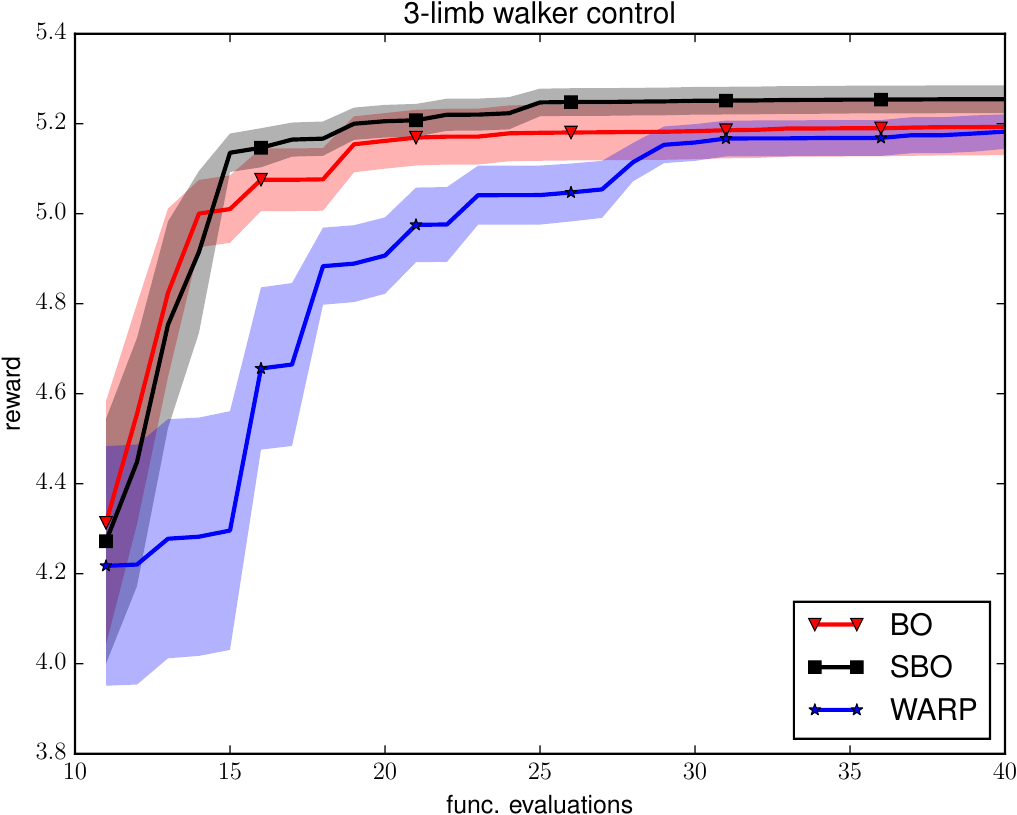}
\includegraphics[width=0.25\linewidth]{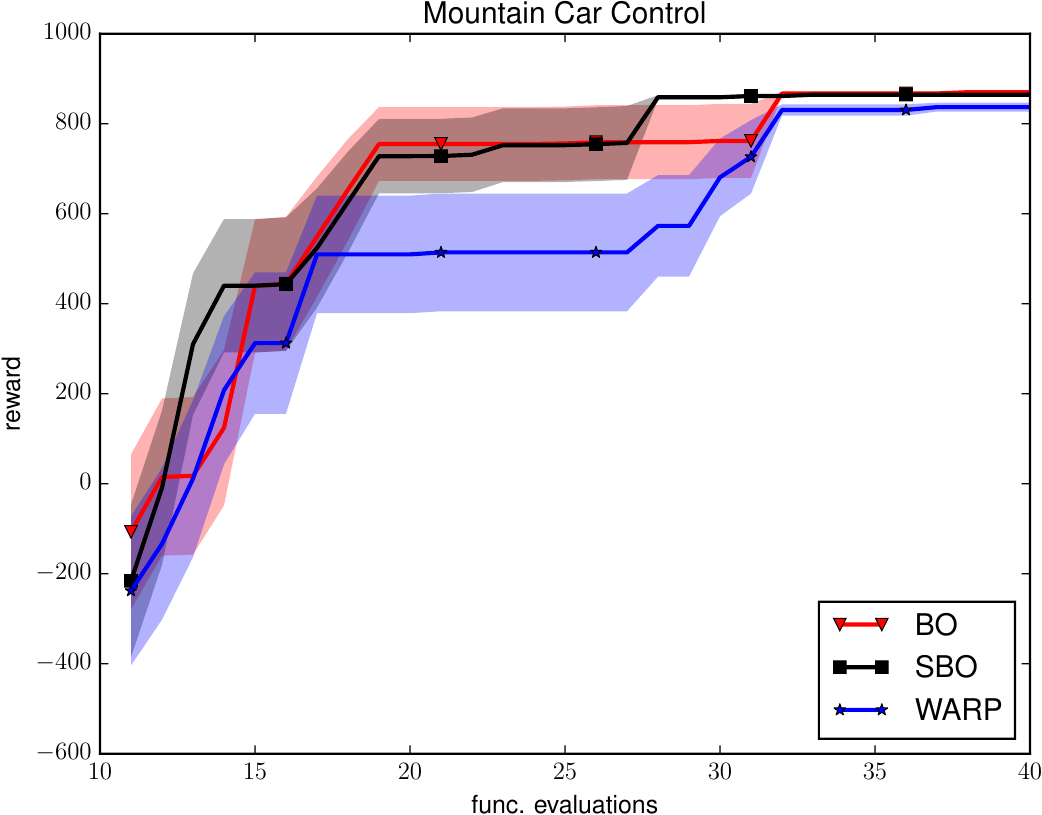}
\includegraphics[width=0.25\linewidth]{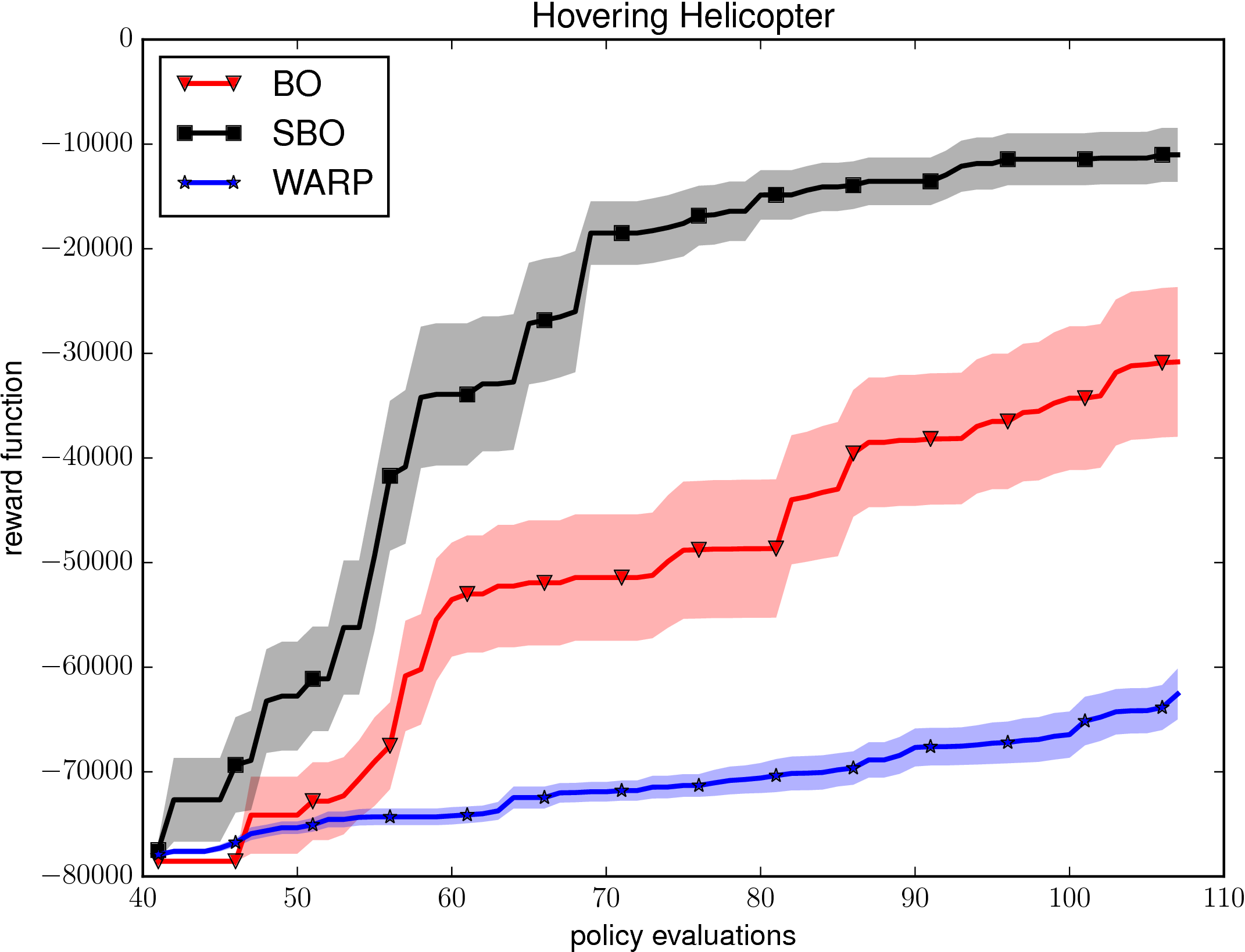}
\caption{Total reward for: a) the three limb walker, b) the mountain car and c) the hovering helicopter control problem. For the first problem, SBO is able to achieve higher reward, while other methods get stuck in a local maxima. For the mountain car, SBO is able to achieve maximum performance in all trials after just 27 policy trials (17 iterations + 10 initial samples). For the helicopter problem, BO and WARP have slow convergence, because many policies results in an early crash, providing almost no information. However, SBO is able to exploit good policies and quickly improve the performance.}
  \label{fig:rlres}
\end{figure*}

We evaluated SBO with several RL/control problems. They all rely on continuous states, actions. We assume the problems are episodic, with a finite time horizon. We have compared our method in three well-known benchmarks with different level of complexity. 

\subsubsection{Walker}
\label{sec:walker} 
The first problem is learning the controller of a three limb robot walker presented in Westervelt et al. \cite{westervelt2007feedback} using their Matlab code. The controller modulates the walking pattern of a simple biped robot. The desired behavior is a fast upright walking pattern, the reward is based on the walking speed with a penalty for not maintaining the upright position. The dynamic controller has 8 continuous parameters. The walker problem was already used as a Bayesian optimization benchmark \cite{NIPS2014_pesshort}.

\subsubsection{Mountain Car}
The second problem is the mountain car problem \cite{sutton1998} based on a Python implementation from Martin H. \cite{MartinFAR}. The state of the system is the car horizontal position. The action is the horizontal acceleration $a \in [-1,1]$. Contrary to the many solutions that discretize both the state and action space, we can directly deal with continuous states and actions. The policy is a simple linear perceptron model inspired by Brochu et al. \cite{Brochu:2010c}. The potentially unbounded policy parameters $\w = \{w_i\}_{i=1}^7$ are computed as $\w=\tan\left(\left(\pi-\epsilon_\pi\right)\w_{01} - \pi/2\right)$ where $\w_{01}$ are the policy parameters bounded in the $[0,1]^7$ space. The term $\epsilon_\pi > 0$ was used to avoid $w_i \rightarrow \infty$.

\subsubsection{Helicopter hovering}
The third problem is the hovering helicopter from the RL-competition \footnote{http://www.rl-competition.org/}. It is one of the most challenging scenarios of the competition, being presented in all the editions since 2008. This problem is based on a simulator of the XCell Tempest aerobatic helicopter. The simulator model was learned based on actual data from the helicopter using apprenticeship learning \cite{Abbeel2006}. The model was used to learn a policy that was later used in the real robot. The simulator included several difficult wind conditions. The state space is 12D (position, orientation, translational velocity and rotational velocity) and the action is 4D (forward-backward cyclic pitch, lateral cyclic pitch, main collective pitch and tail collective pitch). The reward is a quadratic function that penalizes both the state error (inaccuracy) and the action (energy). Each episode is run during 10 seconds (6000 control steps). If the simulator enters a terminal state (crash), a large negative reward is given, corresponding to getting the most negative reward achievable for the remaining time.

We used the weak baseline controller that was included with the helicopter model. This weak controller is a simple linear policy with 12 parameters (weights). In theory, this controller is enough to avoid crashing but is not very robust. We show how this policy can be easily improved with few iterations. In this case, initial exploration of the parameter space is specially important because the number of policies not crashing in few control steps is very small. For most policies, the reward is the most negative reward achievable. Thus, in this case, we have used Sobol sequences for the initial samples of Bayesian optimization. These samples are deterministic, therefore we guarantee that the same number of non-crashing policies are sampled for every trial and every algorithm. We also increased the number of initial points to 40 due to the higher dimensionality.

Figure \ref{fig:rlres} shows the performance for the three limb walker presented, the mountain car and the helicopter problem. In all cases, the results obtained by SBO were more efficient in terms on number of trials and accuracy, with respect to standard BO and WARP. Furthermore, we found that the results of SBO were comparable to those obtained by popular reinforcement learning solvers like SARSA \cite{sutton1998}, but with much less information and prior knowledge about the problem. For the helicopter problem, other solutions found in the literature require a larger number of scenarios/trials to achieve similar performance \cite{koppejan2011neuroevolutionary}.

\subsection{Automatic wing design using CFD simulations}
\label{sec:wing-design-with}

 
The next test is to find the shape of the wing that minimizes the drag while maintaining enough lift. Wing design using computational fluid dynamics (CFD) simulators is also a well known difficult optimization problem due to the chaotic nature of fluid dynamics \cite{forrester2006optimization}. Even though we use a commercial CFD software (XFlow) to simulate the wind tunnel, sample efficiency is mandatory as an average CFD simulation can still take days or months of CPU time.

First, as a common practice in this kind of problems, we assumed a 2D simulation of the fluid along the profile of the wing. This is a reasonable assumption for wings with large aspect ratio (large span compared to the chord), and it considerably reduces the wall time of each simulation from days to hours. For the parametrization of the profile, there are many alternatives based on geometric or manufacturing principles. In our case, we used Bezier curves for their simplicity to generate the corresponding shape. However, note that Bayesian optimization is agnostic of the geometric parametrization and any other parametrization could also be used. The Bezier curve of the wing was based on 7 control points, that is, 14 parameters which were reduced to 5 by adding some physical and manufacturing restrictions.



\begin{figure}
  \centering
\includegraphics[width=0.50\linewidth]{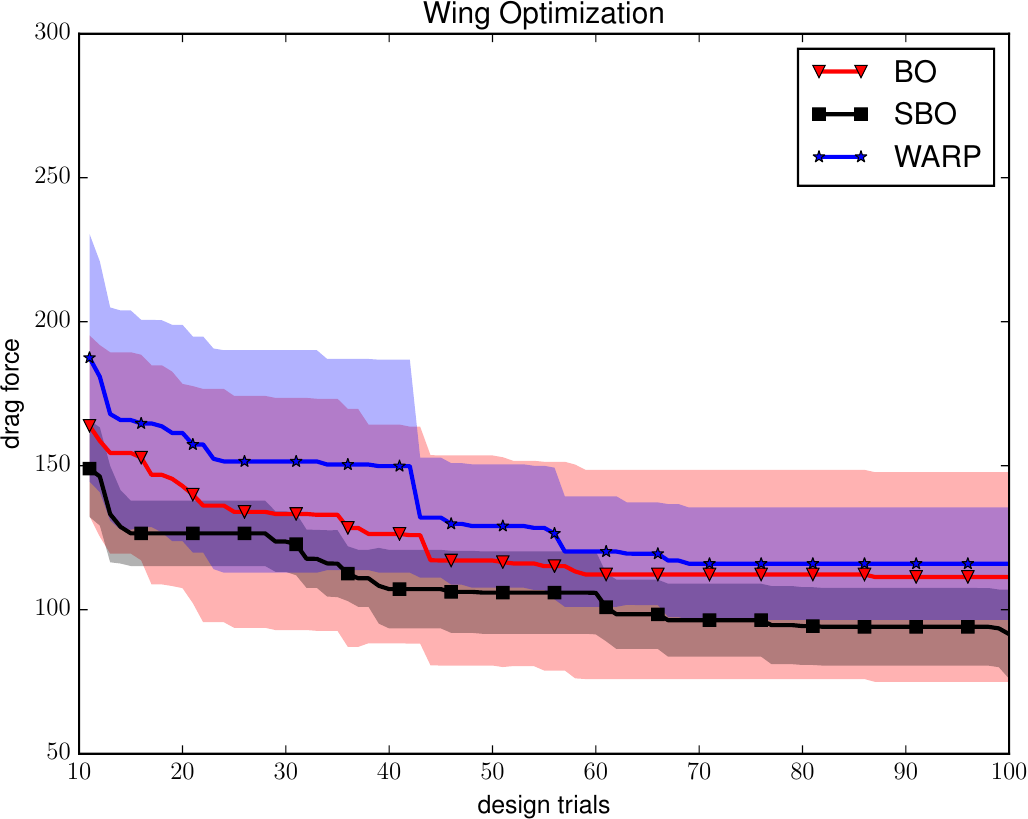}
  \caption{Results for the wing design optimization (10 runs per plot).}
  \label{fig:wingres}
\end{figure}

The problem of minimizing the drag directly is that the best solutions tend to generate flat wings that do not provide enough lift for the plane. As a simple approach, we added a large penalty to the wings without enough lift. We also found that, due to fluid dynamics, the drag value was very chaotic. For example, the flow near the trailing edge can transition from laminar to turbulent regime due to a small change in the wing shape. Thus, the resulting forces are completely different, increasing the drag and reducing the lift. Figure \ref{fig:wingres} shows how both BO and WARP fail to find the optimum wing shape, under these conditions. However, SBO finds a better wing shape and in very few iterations.

\subsection{Computational cost}
\label{sec:time}

The main difference between the three methods (BO, SBO and WARP) in terms of the algorithm is within the kernel function $k(\cdot,\cdot)$, which includes the evaluation of the weights in SBO and the evaluation of the warping function (the cumulative density function of a Beta distribution) in WARP. We found that the time differences between the algorithms were mainly driven by the dimensionality and shape of the posterior distribution of the kernel hyperparameters because MCMC was the main bottleneck. Also, the evaluation of the Beta CDF was more involved and computationally expensive than the evaluation of the Mat{\'e}rn kernel or the Gaussian weights. That extra cost became an important factor as the kernel function is called millions of times for each Bayesian optimization run.

Table \ref{tab:time} shows the average CPU time of the different experiments for the total number of function evaluations. We did not include the helicopter and wing problems because both rely on multiple process synchronization, convoluting the measurements, although the relative computation time were similar to other examples.

\section{Conclusions}

In this paper, we have presented a new algorithm called Spartan Bayesian Optimization (SBO) in the context of \emph{active policy search} for robot control. Our method combines a local and a global kernel in a single adaptive kernel to deal with the exploration/exploitation trade-off and the inherent nonstationarity in the search process during policy search using Bayesian optimization. For nonstationary problems, like robot control problems, the method provides excellent results compared to standard Bayesian optimization and the state of the art method for nonstationarity. Furthermore, SBO also performs well in stationary problems by improving local refinement while retaining global exploration capabilities. We evaluated the algorithm extensively in standard optimization benchmarks and control/reinforcement learning scenarios. Moreover, our contribution can be directly applied to other setups. As an example, we also evaluated SBO in an autonomous wind design problem. The results have shown that SBO increases the convergence speed and reduces the number of samples in many problems. In addition, we have shown how SBO is more efficient in terms of CPU usage than other nonstationary methods for Bayesian optimization. 

\begin{table}
\scriptsize
\centering
  \caption{Average CPU time for the total optimization (in seconds).}
  \label{tab:time}
  \begin{tabular}{|l||r|r|r|r||r|r|}
\hline
\textbf{Time(s)} & \textbf{Gram.} & \textbf{Branin} & \textbf{Hartm.} & \textbf{Michal.}  & \textbf{Walker} & \textbf{MCar}\\
\hline \hline
\#dims & 2  & 2 & 6 & 10 & 8 & 7 \\
\hline
\#evals & 60 & 40 & 70  & 210 & 40 & 40 \\
\hline\hline
  BO    & 120 & 171 & 460 & 8\,360 & 47 & 38  \\
\hline
  SBO   & 2\,481 & 3\,732  & 10\,415 & 225\,313 & 440 & 797 \\
\hline
  WARP  & 13\,929 & 28\,474 & 188\,942 & 4\,445\,854 & 20\,271 & 18\,972 \\
\hline
  \end{tabular}
\end{table}

\bibliographystyle{IEEEtran}
\bibliography{IEEEabrv,../bib/optimization}

\end{document}